\documentclass[letterpaper]{article} 
\usepackage{aaai2027}  
\usepackage[hyphens]{url}  
\usepackage{graphicx} 
\usepackage{natbib}  
\usepackage{caption} 
\usepackage{algorithm}
\usepackage{algorithmic}

\usepackage{newfloat}
\usepackage{amsfonts}
\usepackage{amsmath}
\usepackage{listings}
\DeclareCaptionStyle{ruled}{labelfont=normalfont,labelsep=colon,strut=off} 
\floatstyle{ruled}
\newfloat{listing}{tb}{lst}{}
\floatname{listing}{Listing}

\usepackage{booktabs}

\newcommand{\method}{MPAO}

\title{Many Plans at Once: Diverse Top-Quality Planning in Continuous Time}
\author{
    Mahyar Jahani-nasab,
    Hamid Rezatofighi,
    Mor Vered,
    Buser Say
}
\affiliations{
    Monash University\\
    \{mahyar.jahaninasab, hamid.rezatofighi, mor.vered, buser.say\}@monash.edu
}

\begin{document}

\maketitle

\begin{abstract}
Planners for physical systems usually return a single trajectory, which leaves no alternative when an unmodeled constraint invalidates the chosen route or downstream preferences change. A set of plans computed in advance provides ready-made alternatives and warm-starts for fast adaptation, but only if the plans are both low-cost and mutually distinct. Discrete planning addresses
this with top-quality and diverse planning, but these ideas do not transfer to
continuous time, where every valid trajectory has infinitely many valid
near-copies, and existing trajectory planners converge to one or a few
solutions. We present Many Plans at Once (\method{}), which treats every plan
it finds as a reason to search elsewhere: it samples untrained neural networks
whose outputs meet the boundary conditions by construction, trains corrections
shared by networks from the same distribution, stores every validated plan, and
adapts its sampling to what it has found, which makes it anytime. We define
the goal by bringing the cost bound of top-quality planning and the
distinctness threshold of diverse planning to continuous time. Under equal
budgets against five baselines, \method{} is the only method that solves all
40 instances across four domains. On contact-rich manipulation, it is the only
method with plans within 25\% of the best known cost on one task, and returns
about ten times as many such plans as the strongest baseline on the other two.
\end{abstract}


\section{Introduction}

\begin{figure}[t]
\centering
\includegraphics[width=0.9\columnwidth]{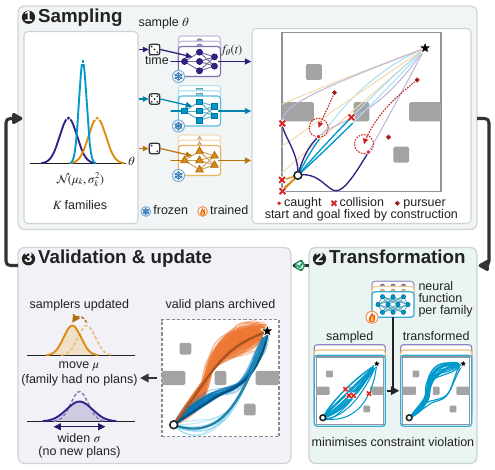}
\caption{\textbf{MPAO in three steps} on real pursuit--evasion data.
\textbf{(1)~Sampling:} Untrained network weights $\mathcal{N}(\mu_k,\sigma_k^2)$ generate initial trajectories with fixed endpoints.
\textbf{(2)~Transformation:} Trained neural functions optimize these trajectories to minimize costs and constraint violations.
\textbf{(3)~Validation \& update:} Valid trajectories are archived and the samplers updated.}
\label{fig:teaser}
\end{figure}

Trajectory planning for physical systems computes a plan that satisfies the system's differential dynamics and
constraints, and typically returns a single trajectory that is optimal, or
locally optimal, with respect to a cost~\cite{betts1998survey, zucker2013chomp,
morari2015predictive}. This paradigm is effective when the environment and objective are fully known, but brittle when new constraints emerge or task preferences shift. If an unexpected obstacle obstructs the chosen route or a secondary preference rules it out, the single plan is unusable and no alternative is available~\cite{otte2016rrtx}. Local feedback control and trajectory deformation can absorb only small deviations around a single nominal path~\cite{tedrake2010lqr}, whereas replanning from scratch across different homotopy classes or contact modes is computationally expensive~\cite{posa2014direct}. Computing a set of plans in advance addresses both limitations: downstream execution or local replanning can fall back on or warm-start from an unobstructed mode~\cite{magdici2016fail}, and an operator can select among qualitatively different options~\cite{nguyen2012generating}. However, such a set is only useful if its plans are both low-cost and mutually distinct: an expensive plan is a poor choice~\cite{katz2020top}, and a plan close to an obstructed one is likely to share the same obstruction~\cite{srivastava2007domain}.

Returning multiple plans is well studied in discrete planning: top-$k$
planning returns the $k$ cheapest plans~\cite{katz2018novel}, top-quality
planning returns all plans within a cost bound~\cite{katz2020top}, and diverse
planning returns plans that differ by a given
distance~\cite{nguyen2012generating, katz2020reshaping}, often by forbidding
each plan found and searching again. These ideas do not transfer directly to
trajectory planning, where a plan is a function of continuous time rather than
a timed sequence of discrete actions, as in PDDL+
planning~\cite{cashmore2020planning}. Shifting a valid path by a fraction of a millimetre yields another valid plan that passes every obstacle on the same side. Consequently, a valid plan has infinitely many valid near-copies. Because these plans cannot be enumerated, forbidding one plan simply leaves all of its near-copies available.

Existing continuous methods fall into three groups, none of which is designed
to return such sets. Trajectory optimizers, whether gradient-based or
sampling-based such as CEM and MPPI, converge to a single local optimum or
mode, and restarting them tends to return the same few
optima~\cite{marti2025multi, kobilarov2012cross, williams2017model}.
Multi-modal methods such as Stein variational MPC maintain several
trajectories, but provide neither cost bounds nor an explicit distinctness
criterion~\cite{lambert2020stein}. Quality-diversity algorithms keep archives
of diverse solutions, but measure diversity over predefined behaviour
descriptors and bound no solution's cost relative to the
best~\cite{cully2015robots}. Because no method targets sets that are both
low-cost and distinct, continuous planning also lacks a measure for them:
benchmarks report success rates, mode counts or
coverage~\cite{chamzas2021motionbenchmaker, cully2015robots}, none of which
combines a cost bound relative to the best plan with a distinctness threshold
on dynamically feasible, constraint-satisfying trajectories.

We address this gap with Many Plans at Once (\method{}), which returns large
sets of distinct, low-cost, validated trajectories in a single run
(Fig.~\ref{fig:teaser}). We formalize the goal as $(q, \delta)$-plan sets (Sec.~\ref{sec:problem}),
bringing the cost bound of top-quality planning and the distinctness threshold
of diverse planning to continuous time, and measure them by their size
$N^\star(q, \delta)$. Unlike optimizers, which stop once they converge,
\method{} treats every plan it finds as a reason to search elsewhere. It samples untrained neural networks whose outputs meet the boundary conditions by construction, and trains a correction shared by networks from the same distribution so that progress on one candidate carries over to others. \method{} then stores every validated plan, replaces most successful candidates with fresh samples, and continuously adapts its sampling distributions to what it has found. Because
stored plans are never discarded, it is anytime. Under equal budgets against
five baselines, \method{} is the only method that solves all 40 instances of
four domains and the only one with any plan within 25\% of the best known cost
on Pick, one of three manipulation tasks (Sec.~\ref{sec:benchmarks}). On the
other two, Push and Slide, it overtakes the strongest baseline's final count
within a tenth of the budget and ends with about ten times as many such plans,
and about 200 times as many within twice that cost.

\section{Related Work}
\label{sec:related}

\paragraph{Continuous planning and trajectory optimization.} Continuous
planners differ from \method{} in what they return.
Gradient-based trajectory optimizers~\cite{zucker2013chomp, betts1998survey}
return one locally optimal trajectory, and restarting
them~\cite{marti2025multi} returns several that often coincide; GPU-parallel
optimizers run many seeds at once and return the
best~\cite{sundaralingam2023curobo}.
Sampling-based planners~\cite{karaman2011sampling} return the best path in
their tree, and sampling-based MPC such as CEM and
MPPI~\cite{kobilarov2012cross, williams2017model} return the trajectory on
which their sampling distribution has concentrated. Stein variational MPC returns
several modes~\cite{lambert2020stein}, but with no bound on their cost and no
explicit criterion for how different they must be; homotopy-aware search
returns paths in distinct topological
classes~\cite{bhattacharya2012topological}, typically without dynamics or
contact. Motion planning has also studied sets of diverse short paths~\cite{voss2015heuristic, green2010toward, knepper2009path} and their survivability when the environment changes~\cite{erickson2009survivability}, though typically for kinematic paths without contact. Separately, multimodal trajectory optimization~\cite{osa2020multimodal} finds several locally optimal trajectories. Forbidding a neighbourhood of each found plan inside a trajectory
optimizer is a natural continuous analogue of forbidding plans, and we evaluate
it as a baseline (Sec.~\ref{sec:setup}). \method{} instead returns a set of plans that are each checked against the dynamics and constraints and pairwise at least $\delta$ apart, so that its plans within any cost bound form a plan set. Finally, it keeps adding to this set after the first plan is found.

\paragraph{Population search and quality-diversity.} Keeping many candidate
solutions at once is the domain of population-based search. Evolutionary
computation maintains diverse populations through niching and restarts, and
quality-diversity algorithms such as MAP-Elites keep archives of elites over
predefined behaviour descriptors~\cite{cully2015robots}. \method{} combines an
archive and restarts with gradient-trained trajectory shifts and boundary
conditions that hold by construction, and measures distinctness by trajectory
distance rather than predefined descriptors.

\paragraph{Neural and learned trajectory generators.} Recent works use neural networks, in
data-driven or data-free form. Data-driven planners, such as diffusion models,
generate diverse trajectories but require demonstration
data~\cite{janner2022planning, carvalho2025motion}; \method{} needs none.
Among data-free approaches, physics-informed neural
networks represent a single solution trajectory per
problem.
\method{} instead optimizes its whole population without training data,
returning a set of distinct, low-cost plans.

\section{Problem Formulation}
\label{sec:problem}

Following~\citet{jahani2026ping}, a continuous planning problem over a horizon
$T > 0$ is a tuple
\[
\Pi = \langle \mathcal{S}, \mathcal{A}, \mathcal{C}, \mathcal{F}, \mathcal{I},
\mathcal{G}, \mathcal{M}, d \rangle
\]
with state space $\mathcal{S}$, action space
$\mathcal{A} \subseteq \mathbb{R}^{n_f}$, hard constraints
$\mathcal{C} : \mathcal{S} \times \mathcal{A} \to \{0, 1\}$, dynamics
$\mathcal{F}$, initial and goal conditions $\mathcal{I}$ and $\mathcal{G}$,
cost $\mathcal{M}$ and, added in this paper, a plan distance $d$. We define
each element below.

\paragraph{Plans, actions and states.} A \emph{plan} is a trajectory
$f \in \mathcal{T} = C^s([0, T], \mathbb{R}^{n_f})$ of the $n_f$ directly
controlled coordinates, such as a robot's position. Its \emph{action} is the
derivative $a_f = f^{(r)}$ of a fixed order $r \le s$, for example an
acceleration ($r = 2$) or a velocity ($r = 1$). The state splits as
$x = (y, z)$. The controlled part $y$ holds the coordinates and their
derivatives below order $r$, which form a chain of integrators driven by the
action, so along a plan $y_f = (f, \dot f, \dots, f^{(r-1)})$. The remaining
part $z$, such as the states of other agents or of manipulated objects,
evolves as $\dot z = \mathcal{F}_z(y, z, a)$, and $z_f$ is obtained by
integrating this equation along the plan from $z_f(0) = z_0$. The dynamics
$\mathcal{F}$ thus consist of the integrator chain for $y$ and of
$\mathcal{F}_z$ for $z$, and a plan determines its full state
$x_f = (y_f, z_f)$. We assume $\mathcal{F}_z$ is continuous and locally
Lipschitz in $z$, and that $z_f$ exists on $[0, T]$ for every plan, so $x_f$
is well defined and unique.

\paragraph{Initial and goal conditions.} $\mathcal{I}$ prescribes $z_0$ and
$f^{(\kappa)}(0) = \bar f_0^{(\kappa)}$ for $\kappa = 0, \dots, s - 1$: orders
$\kappa < r$ set the initial value of $y$, and orders $r \le \kappa < s$ set
the initial action and its derivatives, such as a start from rest.
$\mathcal{G}$ has two parts. The terminal value of the plan is exact:
$f(T) = \bar f_T$. All other terminal requirements hold within
\emph{terminal tolerances}: for $j = 1, \dots, m$, the \emph{normalized goal
residual} $\gamma_j(f) = \lVert e_j(f) \rVert / \epsilon_j$ must satisfy
$\gamma_j(f) \le 1$ for a tolerance $\epsilon_j > 0$, where the residual
$e_j(f)$ is either a derivative error
$f^{(\kappa_j)}(T) - \bar f_T^{(\kappa_j)}$ with $1 \le \kappa_j \le s$, or
the terminal error of a component of $z_f$, such as the position of a
manipulated object. Initial conditions describe the known start of execution
and hold exactly, and the terminal tolerances also cover states that the plan
does not parameterize. We call the exact conditions on the plan,
$f^{(\kappa)}(0) = \bar f_0^{(\kappa)}$ for $\kappa < s$ and
$f(T) = \bar f_T$, its \emph{boundary conditions}.

\paragraph{Cost, validity and distance.} The cost
$\mathcal{M}(f) = \int_0^T c(x_f(t), a_f(t), \dot a_f(t))\,dt$ integrates a
nonnegative running cost $c$, with $s \ge r + 1$ whenever $c$ depends on
$\dot a_f$. A plan is \emph{valid}, $f \in \mathcal{T}_{\mathrm{valid}}$, if
it meets $\mathcal{I}$ and $\mathcal{G}$ and
$\mathcal{C}(x_f(t), a_f(t)) = 1$ for all $t \in [0, T]$. We assume
$\mathcal{T}_{\mathrm{valid}} \neq \emptyset$ and call
$\mathcal{M}^{\mathrm{ref}} = \inf_{f \in \mathcal{T}_{\mathrm{valid}}}
\mathcal{M}(f) > 0$ the \emph{optimal cost}. To compare plans, let
$p_f(t) \in \mathbb{R}^{m'}$ collect the plan's \emph{task-relevant
positions} at time $t$, such as those of the robot and of manipulated objects,
and let $t_1, \dots, t_P$ be $P$ uniformly spaced \emph{nodes} in $[0, T]$.
Stacking the sampled positions gives
$\Phi(f) = (p_f(t_1), \dots, p_f(t_P)) \in \mathbb{R}^{P m'}$, and the
\emph{plan distance}
$d(f, f') = \lVert \Phi(f) - \Phi(f') \rVert_2 / (L \sqrt{P})$ is the
root-mean-square distance between the task-relevant positions of the two
plans at the nodes, normalized by a domain scale $L > 0$.

\paragraph{Multi-plan planning.} For a \emph{cost slack} $q \ge 0$, let
$\mathcal{T}_q = \{ f \in \mathcal{T}_{\mathrm{valid}} :
\mathcal{M}(f) \le (1+q)\,\mathcal{M}^{\mathrm{ref}} \}$, which is nonempty
for every $q > 0$, and for $q = 0$ if the infimum is attained.

\noindent\textbf{Definition 1} ($(q, \delta)$-plan set)\textbf{.} For a cost
slack $q \ge 0$ and a \emph{distinctness threshold} $\delta > 0$, a
\emph{$(q, \delta)$-plan set} is a finite set
$\mathcal{P} \subseteq \mathcal{T}_q$ with $d(f, f') \ge \delta$ for all
distinct $f, f' \in \mathcal{P}$.

Because $d$ is only a pseudometric, distinct plans can be at distance zero;
since $\delta > 0$, at most one of them can belong to a $(q, \delta)$-plan
set. Multi-plan planning seeks a $(q, \delta)$-plan set of maximum size.

\noindent\textbf{Proposition 1.} \textit{Assume that, for every valid plan and
every node, each task-relevant position vector lies in a fixed ball of radius
$B$ in $\mathbb{R}^{m'}$. Then every $(q, \delta)$-plan set has at most
$(1 + 2B/(\delta L))^{P m'}$ plans, and a maximum one exists whenever
$\mathcal{T}_q \neq \emptyset$.}

\noindent\textit{Proof.} Under the assumption, every $\Phi(f)$ lies in a ball
of radius $B\sqrt{P}$ in $\mathbb{R}^{P m'}$, and the points $\Phi(f)$ of a
$(q, \delta)$-plan set are pairwise at least $\delta L \sqrt{P}$ apart. Balls
of radius $\delta L \sqrt{P}/2$ around them are disjoint and lie in a ball of
radius $(B + \delta L/2)\sqrt{P}$, and comparing volumes gives the bound. If
$\mathcal{T}_q \neq \emptyset$, every singleton
$\{f\} \subseteq \mathcal{T}_q$ is a $(q, \delta)$-plan set, so the attainable
sizes form a nonempty, bounded set of integers, which has a maximum.
$\square$

The assumption holds whenever every valid plan keeps its task-relevant
positions in a bounded region: for example, when $\mathcal{C}$ confines them
directly, or when it bounds the actions, so that the positions cannot leave a
bounded region over the finite horizon. In our domains, $\mathcal{C}$ keeps
the robot inside a bounded region, and objects move only through contact with
it (Sec.~\ref{sec:benchmarks}). If $\mathcal{T}_q$ is finite and its plans
are pairwise at least $\delta$ apart, as in discrete planning, the maximum
$(q, \delta)$-plan set is $\mathcal{T}_q$ itself, recovering top-quality
planning~\cite{katz2020top}.

\section{Method}
\label{sec:method}

\begin{figure*}[t]
  \centering
  \includegraphics[width=\textwidth]{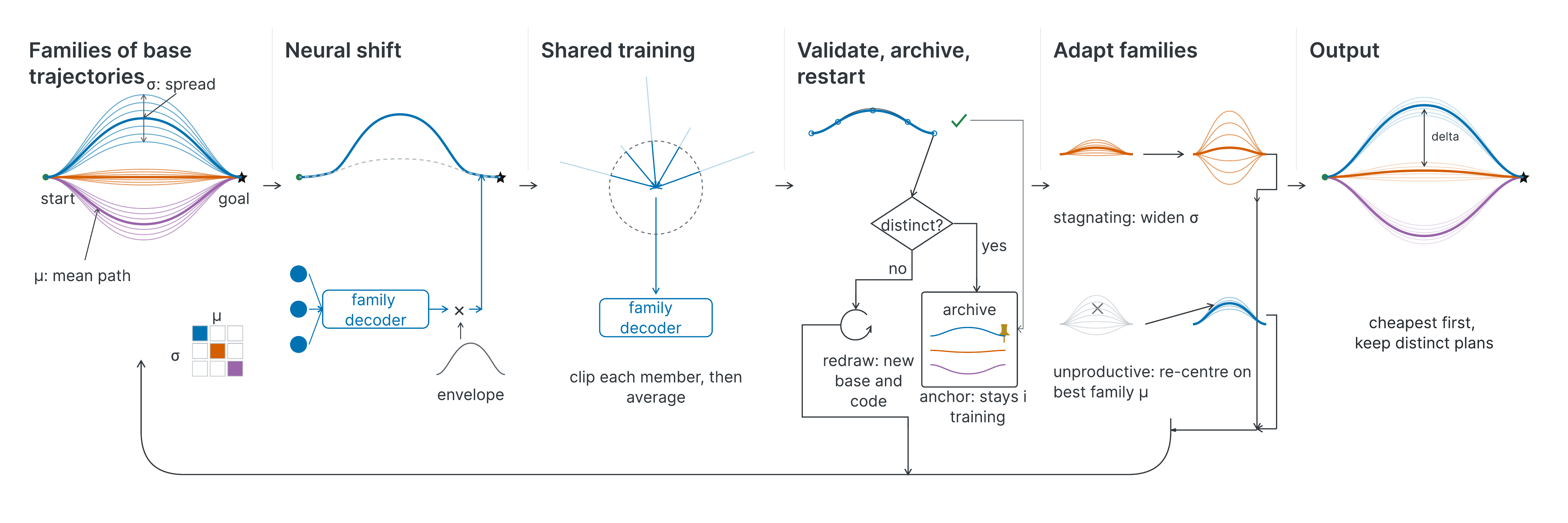}
  \caption{\textbf{Overview of \method{}.} (1)~Families sample base
  trajectories and probe them; (2)~family-shared decoders shift them; (3)~the shifts are
  trained with per-member gradient normalization; (4)~validated plans are
  archived and their slots anchored or redrawn, and slots that stay invalid are
  restarted; (5)~families widen or are replaced; (6)~the archive yields
  distinct, low-cost plans.}
  \label{fig:method_overview}
\end{figure*}

\method{} maintains a population of $N$ \emph{slots}, divided among $K$
\emph{families} (Fig.~\ref{fig:method_overview}). Each slot holds one
\emph{member}: a candidate plan formed by a base trajectory, sampled from its
family's distribution and satisfying the boundary conditions by construction
(Sec.~\ref{sec:base}), plus a shift produced by a decoder shared within
the family (Sec.~\ref{sec:shift}). The shifts are trained by gradient-based
optimization on a loss that penalizes violations of $\mathcal{C}$ and
$\mathcal{G}$ (Sec.~\ref{sec:training}). Every $C$ training epochs, the
members are checked against $\mathcal{C}$ and $\mathcal{G}$; validated plans
are archived, and a slot is redrawn with a new member after a success
(unless it is kept as an anchor) or after $R$ unsuccessful checks, so that no slot keeps an unsuccessful member
indefinitely (Sec.~\ref{sec:archive}). Families that stop finding new
plans widen, and families that find none are replaced
(Sec.~\ref{sec:adapt}).

\subsection{Families of Base Trajectories}
\label{sec:base}

Each base trajectory satisfies the boundary conditions
(Sec.~\ref{sec:problem}) by construction. With $\tau = t/T$,
\begin{equation}
\label{eq:ansatz}
\begin{split}
f_\theta(t) = {} & \sum_{\kappa=0}^{s-1} \frac{t^\kappa}{\kappa!}\, \bar f_0^{(\kappa)}
+ \tau^s \Bigl(\bar f_T - \sum_{\kappa=0}^{s-1} \frac{T^\kappa}{\kappa!}\, \bar f_0^{(\kappa)}\Bigr) \\
& + t^s (t - T)\, \psi_\theta(t),
\end{split}
\end{equation}
where $\psi_\theta : [0, T] \to \mathbb{R}^{n_f}$ is a neural residual with
parameters $\theta$. The factor $t^s$ leaves the prescribed initial
derivatives unchanged, and the factor $t - T$ makes the residual vanish at
$t = T$, so $f_\theta^{(\kappa)}(0) = \bar f_0^{(\kappa)}$ for $\kappa < s$
and $f_\theta(T) = \bar f_T$ for every $\theta$, as in trial-function
methods~\cite{lagaris1998artificial, mortari2017theory}. The residual is a
two-layer network, $\psi_\theta(t) = W_2\, \phi(W_1 t + b_1) + b_2$, with a
nonlinear activation $\phi$.

\paragraph{Families.} Each family $k$ draws the base parameters
$\theta = (W_1, b_1, W_2, b_2)$ of its members by sampling every entry
independently from a Gaussian $\mathcal{N}(\mu_k, \sigma_k^2)$; these
parameters are never trained. The mean $\mu_k$ fixes the family's
\emph{center trajectory}, the base trajectory obtained when every entry of
$\theta$ equals $\mu_k$ (that is, when $\sigma_k = 0$), and the spread
$\sigma_k$ sets how far sampled members deviate from it. Because the Gaussian
gives positive probability to every open set of parameters, a family can
sample trajectories arbitrarily close to any base trajectory that the network
can represent. Initializing the $K$ families on a grid of
$(\mu_k, \sigma_k)$ values covers several center trajectories and spreads.

\subsection{Neural Shift}
\label{sec:shift}

Each family $k$ has one decoder network $D_{\omega_k}$, with weights
$\omega_k$, shared by its members, and each member $i$ has its own latent code
$\zeta_i$. The plan of member $i$, in family $k(i)$, adds a shift to its base
trajectory:
\begin{equation}
\begin{split}
\tilde f_i(t) &= f_{\theta_i}(t) + w_{\text{env}}(\tau)\, D_{\omega_{k(i)}}(t; \zeta_i),\\
w_{\text{env}}(\tau) &= c_s \tau^s (1 - \tau),
\end{split}
\label{eq:shift}
\end{equation}
where the envelope $w_{\text{env}}$ is scaled by $c_s$ so that its peak value
is one. The envelope vanishes to order $s$ at $t = 0$ and to first order at
$t = T$, so the shift changes neither the initial derivatives nor the
terminal value, and $\tilde f_i$ satisfies the boundary conditions for any
decoder weights and code. The terminal tolerances of $\mathcal{G}$ are
enforced by penalties in the training loss (Sec.~\ref{sec:training}).
Because the decoder is shared, improvements learned for one member can carry
over to the rest of its family, while each member's code lets its plan differ
from the others.

\paragraph{Per-member gradient normalization.} So that no single member
dominates a shared decoder, the gradient
$g_i = \nabla_{(\omega_{k(i)},\, \zeta_i)}\, \ell_i$ of each member's training
loss $\ell_i$ (Eq.~\ref{eq:loss}) is clipped to norm at most $\nu$ before the
decoder parts are averaged over the set $F_k$ of members of family $k$:
\begin{equation}
\hat g_i = g_i \min\bigl(1, \nu / \lVert g_i \rVert\bigr),
\qquad
\bar g_k = \frac{1}{|F_k|} \sum_{i \in F_k} \hat g_{i,\omega},
\label{eq:pmn}
\end{equation}
where $\hat g_{i,\omega}$ is the decoder part of $\hat g_i$. Adam updates each
decoder with $\bar g_k$ and each code with its own clipped gradient.

\subsection{Validate, Archive, Restart}
\label{sec:archive}

Every $C$ epochs, each member that is not an anchor is checked in three stages. The first is a node-wise check, which requires $\mathcal{C}(x_{\tilde f_i}(t_p), a_{\tilde f_i}(t_p)) = 1$ at the $P$ nodes and $\gamma_j(\tilde f_i) \le 1$ for all $j$. The second is a replay check, which re-integrates $x_{\tilde f_i}$ on a finer grid, holding the action constant over each control step, and evaluates the constraints again. The third is a certificate, which bounds every constraint on all of $[0, T]$ by interval arithmetic and requires the bounds to rule out any violation. Plans that pass all three stages are \emph{validated} and added to an archive ordered by $\mathcal{M}$. A validated plan is considered \emph{distinct} if its distance $d$ to every previously archived plan is at least $\delta$. Each member then follows one of three rules:
\begin{itemize}
  \item \textbf{Anchor.} If its plan is distinct and its family holds fewer
  than $A_{\max}$ anchors, its base trajectory and code are frozen, and its
  loss stays in the family's objective, keeping the shared decoder near a
  known feasible plan.
  \item \textbf{Redraw.} If its plan is a near-duplicate, or its family holds
  $A_{\max}$ anchors, it receives a new base trajectory from its family's
  current distribution and a fresh code.
  \item \textbf{Restart.} If it is not an anchor and has not been validated
  within $R$ checks of being drawn, it is redrawn in the same way.
\end{itemize}
Other members continue training. Every new base trajectory, whether drawn at initialization or by any rule here or in Sec.~\ref{sec:adapt}, is also \emph{probed}: it is checked at once, without a shift, in the same three stages and archived if it passes.

\paragraph{Restarts and completeness.} The restart mechanism ensures continuous resampling in every slot that is not an anchor, and the probe checks every draw before training shifts it. Provided that each drawn base trajectory passes the check with a probability of at least $p > 0$, the probability of failing to find a plan decays geometrically with the number of draws, whatever the training does (see the Appendix for proofs of completeness and soundness).

\subsection{Adapt Families}
\label{sec:adapt}

\paragraph{Widening.} If a family contains at least one anchor but fails to add a new, distinct plan for $W$ consecutive checks, it widens its search space. It increases its spread by setting $\sigma_k \leftarrow \min(\rho\,\sigma_k, \sigma_{\max})$ and redraws all non-anchor members. The family's anchors, shared decoder, and center $\mu_k$ are preserved, ensuring its center trajectory remains unchanged. The family must then wait another $W$ checks before it is eligible to widen again.

\paragraph{Replacement.} At 25\% and 50\% of the training budget, every family
without an archived plan takes the center $\mu$ of the family with the most
distinct archived plans (ties broken by the cheapest archived plan; if no plan
exists yet, by the smallest constraint violation), resets its spread to
$\sigma_{\mathrm{rep}}$, and redraws its decoder and members.

\subsection{Shared Training}
\label{sec:training}

The constraints $\mathcal{C}$ and the terminal tolerances of $\mathcal{G}$,
which the boundary conditions do not guarantee, are penalized. Each member $i$
minimizes
\begin{equation}
\ell_i = \alpha_C \ell_i^{C} + \alpha_G \ell_i^{G} + \alpha_M \frac{\mathcal{M}(\tilde f_i)}{\bar{\mathcal{M}}_0} + \alpha_R \ell_i^{R},
\label{eq:loss}
\end{equation}
where $\ell_i^{C} = \frac{1}{P}\sum_p v_{ip} + \max_p v_{ip}$ penalizes the mean
and peak violation $v_{ip} \ge 0$ of $\mathcal{C}$ at node $t_p$;
$\ell_i^{G}$ applies a Huber loss to the normalized goal residuals
$\gamma_j(\tilde f_i)$; $\bar{\mathcal{M}}_0$ is the mean cost of the initial
population; and $\ell_i^{R}$
penalizes the shift magnitude. Decoders and codes are optimized with Adam and
a cosine learning-rate schedule.

\subsection{Output and Evaluation Measure}
\label{sec:output}

\method{} filters its archive using a cheapest-first $\delta$-packing strategy before returning the final plans. It evaluates plans in order of increasing cost, keeping a plan only if its distance $d$ to all previously kept plans is at least $\delta$. Consequently, any discarded plan is always within a distance of $\delta$ from a kept plan of equal or lower cost. Because the archive only grows over time (its capacity is never reached in our experiments), the size of its largest $(q, \delta)$-plan set, for a fixed reference cost, never decreases, making \method{} an anytime algorithm (the greedy packing itself can still lose a plan when a new, cheaper plan displaces two kept ones). To compare different methods, we define $\mathcal{M}^\star$ as the lowest validated plan cost found for a given instance across all evaluated methods and seeds (the reference cost $\mathcal{M}^\circ$ of the Appendix). Because $\mathcal{M}^\star \ge \mathcal{M}^{\mathrm{ref}}$, evaluating a set against $\mathcal{M}^\star$ effectively produces a $(q^\circ, \delta)$-plan set, where $q^\circ = (1+q)\,\mathcal{M}^\star/\mathcal{M}^{\mathrm{ref}} - 1 \ge q$.

Our evaluation metric, $N^\star(q, \delta)$, counts the number of packed plans with a cost up to $(1+q)\,\mathcal{M}^\star$. This metric is computed identically across all methods. Since the packing strategy is greedy, $N^\star(q, \delta)$ serves as a lower bound on the maximum possible size of such a set within a method's output. By the triangle inequality, it is also at least the size of any set of the output's plans within the bound that are pairwise at least $2\delta$ apart.

\section{Results and Discussion}
\label{sec:evaluation}
\subsection{Benchmarks}
\label{sec:benchmarks}

We evaluate \method{} on four domains: pursuit--evasion and three manipulation tasks.
Each domain has an instance generator with randomized parameters. We evaluate 10 test instances per domain, alongside two separate held-out instances used specifically to design \method{} and tune the baselines. The full models, constants and generators are in our code
repository.

\paragraph{Pursuit--evasion.} An evader must reach a goal through obstacles
while $n_p$ pursuers chase it. The plan $f$ is the evader's position
$p_e \in \mathbb{R}^2$, and its action is the acceleration $a_f = \ddot p_e$
($r = 2$), bounded by $\lVert a_f \rVert_2 \le a_{\max}$. The remaining state
$z$ holds the pursuers' positions and velocities. Each pursuer $j$ follows
damped pure pursuit with thrust $a_p$, drag $c_d$ and a small smoothing
constant $\eta > 0$:
\begin{equation}
\ddot p_j = \frac{a_p\, (p_e - p_j)}{\sqrt{\lVert p_e - p_j \rVert_2^2 + \eta^2}}
- c_d\, \dot p_j.
\label{eq:pursuit}
\end{equation}
All agents start at rest ($s = 2$).
\textbf{Constraints:} the evader stays in the workspace, a bounded arena, at least
$d_{\min}$ from every obstacle and at least $R_{\mathrm{cap}}$ from every
pursuer.
\textbf{Goal:} $p_e(T) = p_{\mathrm{goal}}$, with terminal velocity
$\lVert \dot p_e(T) \rVert_\infty \le \epsilon_v$.
\textbf{Cost:} $c = \lVert a_f \rVert_2^2
+ w_p \sum_j e^{-\lVert p_e - p_j \rVert_2}$, trading effort against
proximity to the pursuers.
\textbf{Distance:} $\Phi$ samples the evader's position.
Because obstacles can be passed on either side and pursuers evaded in several ways, this domain tests whether a planner finds several different strategies rather than converging to a single plan.

\paragraph{Push, Slide and Pick.} These follow the Fetch tasks of~\citet{plappert2018multigoal}, where a robot arm with a two-fingered gripper moves a block on a table. However, to ensure every method plans against the exact same model, we use our own continuous-time dynamics instead of MuJoCo. The plan
is the commanded grip-site position and gripper aperture,
$f = (p_{\mathrm{cmd}}, \xi) \in \mathbb{R}^4$, and its action is their
velocity ($r = 1$), bounded by $\lVert a_f \rVert_\infty \le a_{\max}$. The
remaining state $z$ holds the joint angles $\mathbf{q}$ and the block's
position $p_o$ and velocity. The joints track the command through damped
differential inverse kinematics, and the block moves through compliant contact
with the gripper and the table. The arm, the block and the command start at
rest ($s = 2$).
\textbf{Constraints:} limits on the workspace (the bounded region the grip site may occupy), the aperture, the
joints' positions, velocities and torques, the tracking error and the
contact forces; the block stays above the table.
\textbf{Goal:} the final command exactly; within tolerances, the block
at its target $p^\star$ and the gripper at rest.
\textbf{Cost:} $c = w_1 \lVert a_f \rVert_2^2
+ w_2 \lVert \dot a_f \rVert_2^2 + w_3 \lVert \ddot{\mathbf{q}} \rVert_2^2$.
\textbf{Distance:} $\Phi$ samples the grip site and the block.
The tasks differ in their dynamics. In \textit{Push}, the target is on the
table. In \textit{Slide}, it lies beyond the arm's reach on a low-friction
surface, so the block must slide after release. In \textit{Pick}, it lies above
the table, so the block must be grasped, with grip forces limited by a
friction cone.

\subsection{Baselines and Setup}
\label{sec:setup}

Plans are distinct at $\delta_1 = 0.0137$ or $\delta_2 = 2\delta_1$, an RMS separation of 1 or 2\,cm on manipulation ($L \approx 0.73$\,m) and of 0.19 or 0.39\,m on pursuit--evasion ($L \approx 14.2$\,m).
We compare against five baselines; forbid-iterative IPOPT and SVMPC are
designed to return several distinct plans.
\textbf{IPOPT}~\cite{wachter2006implementation} solves each problem by direct
multiple shooting in CasADi~\cite{andersson2019casadi}, either from many
starts (1000 on pursuit--evasion, 10 on manipulation) or
forbid-iteratively~\cite{katz2018novel, katz2020top, katz2020reshaping},
re-solving with each new plan constrained to lie at least $\delta_1$ from all
earlier ones. We report the better variant per domain, by
$N^\star(0.25, \delta_1)$.
\textbf{CEM}~\cite{kobilarov2012cross, pinneri2021sample} refits a Gaussian
sampling distribution to its best samples, restarting periodically.
\textbf{MPPI}~\cite{williams2017model, williams2018information} uses smooth,
kernel-interpolated noise and periodic restarts; with
independent per-step noise, it found no validated plan in tuning.
\textbf{SVMPC}~\cite{lambert2020stein} moves a set of trajectory particles by
Stein variational gradient descent, with a kernel that repels them from each
other; we use the implementation of~\citet{barcelos2021dual}.
\textbf{PING}~\cite{jahani2026ping} samples candidate plans in parallel and
refines one of them by physics-informed training. All methods plan using the same model, and each plan is validated through identical node-wise checks and replays. For the baselines, plans are validated at checkpoints during their run. All validated plans across all methods are pooled and packed identically to \method{}. The baselines were tuned on the held-out instances following a pre-registered protocol (selected by $N^\star(0.25, \delta_1)$). \method{} uses $N = 1000$ slots in $K = 9$ families, with initial $(\mu_k, \sigma_k) \in \{-0.3, 0, 0.3\} \times \{0.08, 0.12, 0.18\}$ and all thresholds fixed on the held-out instances. Finally, \method{}, CEM, MPPI, SVMPC, and PING run on a single laptop GPU (NVIDIA RTX 1000 Ada). IPOPT runs on 10--14 parallel CPU workers instead, because its interior-point iterations rely on sparse direct linear solvers optimized for CPUs. However, it still receives the exact same wall-clock budget as every other method.

\subsection{Diverse Top-Quality Planning Across Domains}
\label{sec:main-results}

\begin{table*}[t]
\centering
\small 
\setlength{\tabcolsep}{4pt} 
\caption{\textbf{Diverse plans within the cost bound, by domain and method.}
Mean over 10 test instances. Solved: instances with at least one validated
plan. Bold: best per domain, with its margin over the runner-up.}
\label{tab:main}
\begin{tabular}{llcrrrrr}
\toprule
 & & & \multicolumn{3}{c}{$N^\star(q, \delta)$} & & \\
\cmidrule(lr){4-6}
Domain & Method & Solved & $q{=}0.10,\ \delta_1$ & $q{=}0.25,\ \delta_1$ & $q{=}0.25,\ \delta_2$ & Best$/\mathcal{M}^\star$ & First plan [s] \\
\midrule
Pick
 & \textbf{\method{}}  & \textbf{10/10} & \textbf{0.7} & \textbf{13.5} & \textbf{8.5} & \textbf{1.082}$^{\uparrow 75.2\%}$ & \textbf{16.9}$^{\downarrow 89.6\%}$ \\
 & PING  & 0/10 & 0.0 & 0.0 & 0.0 & -- & -- \\
 & CEM  & 0/10 & 0.0 & 0.0 & 0.0 & -- & -- \\
 & IPOPT  & 0/10 & 0.0 & 0.0 & 0.0 & -- & -- \\
 & MPPI  & \textbf{10/10} & 0.0 & 0.0 & 0.0 & 4.368 & 162.8 \\
 & SVMPC   & \textbf{10/10} & 0.0 & 0.0 & 0.0 & 44.817 & 324.9 \\
\midrule
Push
 & \textbf{\method{}}  & \textbf{10/10} & \textbf{1.1}$^{\uparrow 175\%}$ & \textbf{6.6}$^{\uparrow 1220\%}$ & \textbf{3.0}$^{\uparrow 500\%}$ & \textbf{1.071}$^{\uparrow 7.0\%}$ & \textbf{20.1}$^{\downarrow 42.1\%}$ \\ 
 & PING  & 0/10 & 0.0 & 0.0 & 0.0 & -- & -- \\
 & IPOPT   & 6/10 & 0.4 & 0.5 & 0.5 & 1.151 & 131.7 \\
 & CEM  & 7/10 & 0.0 & 0.0 & 0.0 & 4.681 & 34.7 \\
 & MPPI  & \textbf{10/10} & 0.0 & 0.0 & 0.0 & 7.882 & 83.6 \\
 & SVMPC & 1/10 & 0.0 & 0.0 & 0.0 & 64.470 & 794.6 \\
\midrule
Slide
 & \textbf{\method{}}  & \textbf{10/10} & \textbf{1.1}$^{\uparrow 175\%}$ & \textbf{3.9}$^{\uparrow 875\%}$ & \textbf{2.9}$^{\uparrow 625\%}$ & \textbf{1.053}$^{\uparrow 1.0\%}$ & \textbf{18.2}$^{\downarrow 60.9\%}$ \\ 
 & PING  & 0/10 & 0.0 & 0.0 & 0.0 & -- & -- \\
 & IPOPT & 7/10 & 0.4 & 0.4 & 0.4 & 1.064 & 313.6 \\
 & CEM  & \textbf{10/10} & 0.0 & 0.0 & 0.0 & 3.191 & 46.5 \\
 & MPPI  & \textbf{10/10} & 0.0 & 0.0 & 0.0 & 4.744 & 162.0 \\
 & SVMPC  & 0/10 & 0.0 & 0.0 & 0.0 & -- & -- \\
\midrule
Pursuit--evasion
 & \textbf{\method{}}  & \textbf{10/10} & 7.7 & 39.9 & \textbf{8.3}$^{\uparrow 33.9\%}$ & 1.050 & 15.4 \\
 & PING  & 0/10 & 0.0 & 0.0 & 0.0 & -- & -- \\
 & IPOPT  & \textbf{10/10} & \textbf{27.5}$^{\uparrow 248.1\%}$ & \textbf{40.3}$^{\uparrow 1.0\%}$ & 6.2 & \textbf{1.000}$^{\uparrow 4.8\%}$ & \textbf{4.6}$^{\downarrow 57.0\%}$ \\
 & CEM & 9/10 & 0.0 & 0.6 & 0.6 & 1.232 & 10.7 \\
 & MPPI   & 9/10 & 7.9 & 25.8 & 3.6 & 1.061 & 12.8 \\
 & SVMPC & \textbf{10/10} & 0.0 & 0.0 & 0.0 & 1.628 & 37.5 \\
\bottomrule
\end{tabular}
\end{table*}

Table~\ref{tab:main} reports solved instances, $N^\star(q, \delta)$ across two cost bounds and two distinctness thresholds, best plan cost, and time to first plan by method and domain.

\paragraph{\method{} is the only method competitive on every domain.} It solves
all 40 instances and is the only method that returns plans within 25\% of the
best known cost $\mathcal{M}^\star$ on every domain. On the three manipulation
domains, it matches the highest solve rate and outperforms every baseline on
every other measure: it returns the most distinct plans, finds the cheapest
plan, and finds its first plan soonest.

\paragraph{Contact-rich manipulation.} \method{} achieves its largest lead on these domains. Within 25\% of $\mathcal{M}^\star$, \method{} returns an average of 13.5 distinct plans per instance on Pick, 6.6 on Push, and 3.9 on Slide. In contrast, the strongest baseline (IPOPT) manages only 0.5 on Push and 0.4 on Slide, and fails to find any on Pick. Consequently, \method{} generates plan sets that are 13 times larger on Push and 10 times larger on Slide (or 11 and 8 times larger when averaging \method{} across three seeds). Furthermore, it is the only method that finds any plan within this bound on Pick. These leads are statistically significant: across all three domains, the 95\% bootstrap interval for \method{}'s advantage remains strictly above zero at both $\delta_1$ and the stricter $\delta_2$. At the
tighter bound $q = 0.10$, \method{} leads in the mean everywhere, but so few
plans qualify that the difference is clear only on Pick. We also grouped plans by contact strategy using predetermined features (e.g., how the gripper first meets the block, push count, pre-grasp pushing, and release location). Under this classification, \method{}'s plans span 3.2, 3.0, and 1.4 different strategies per instance on Pick, Push, and Slide, compared to at most 0.5 for any baseline. While excluding plans near class boundaries lowers the absolute counts, \method{} maintains this commanding lead. Finally, high solve rates are
deceptive here: MPPI solves every manipulation instance, but its cheapest
plans cost 4.4 to 7.9 times $\mathcal{M}^\star$, so none fall within the
bound.

\paragraph{Pursuit--evasion.} Without contact, local optimization works well
and the strongest baselines are competitive. IPOPT matches \method{} at $\delta_1$ (40.3 vs.\ 39.9 plans, or 40.6 for \method{} averaged over three seeds). Furthermore, IPOPT and MPPI find more plans very close to $\mathcal{M}^\star$ (27.5 and 7.9 vs.\ 7.7 at $q = 0.10$), and IPOPT finds its first plan sooner (4.6 vs.\ 15.4\,s). When plans must differ more, \method{} leads: at $\delta_2$, it
returns more distinct plans than IPOPT on 7 of 10 instances (8.3 vs.\ 6.2 on
average). Much of IPOPT's count also comes from timing rather than path: under
the discrete Fr\'echet distance, which ignores when a path reaches each point,
IPOPT's $N^\star(0.25, \delta_1)$ falls from 40.3 to 11.6, while \method{}
keeps 23.6 and leads on 7 of 10 instances (MPPI keeps 22.9). Homotopy classes
around the static obstacles give a coarser view: the plans fall into two or
three classes per instance, and \method{}'s archive contains every class
IPOPT finds, plus one more on one instance. Within $1.25\,\mathcal{M}^\star$,
however, IPOPT reaches more of them (1.7 vs.\ 1.3 per instance), because
\method{}'s plans on the second route cost more.

\subsection{Robustness Across Cost Bounds, Seeds and Thresholds}
\label{sec:robustness}

\begin{figure}[t]
\centering
\includegraphics[width=0.8\columnwidth]{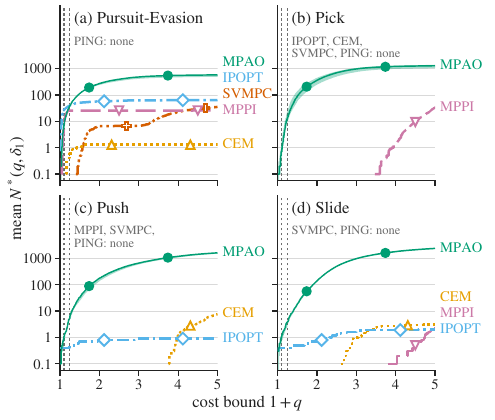}
\caption{\textbf{Distinct plans as the cost bound is relaxed.} Mean
$N^\star(q, \delta_1)$ over the 10 test instances against the cost bound
$1+q$ (log scale).}
\label{fig:robustness}
\end{figure}

Fig.~\ref{fig:robustness} varies the bound from $\mathcal{M}^\star$ to
$5\,\mathcal{M}^\star$. On manipulation, \method{} leads every baseline at every bound above $1.05\,\mathcal{M}^\star$, eventually exceeding 1{,}000 plans per instance on all three domains. In contrast, the baselines' first plans only appear at several times $\mathcal{M}^\star$ (e.g., MPPI starts at $3.47\,\mathcal{M}^\star$ on Pick, and CEM at $2.6\,\mathcal{M}^\star$ on Slide), and IPOPT never exceeds 2 plans per instance at any bound. Within $2\,\mathcal{M}^\star$, \method{} is the only
method with plans on Pick and returns about 200 times as many as the best
baseline on Push and Slide. On pursuit--evasion, IPOPT leads up to the $1.25\,\mathcal{M}^\star$ bound. Beyond this point, however, the baselines saturate while \method{}'s margin of victory grows to 4.8 times at $2\,\mathcal{M}^\star$ and 9.1 times at $5\,\mathcal{M}^\star$. These conclusions hold for every seed: \method{}'s weakest
seed leads every baseline above $1.1\,\mathcal{M}^\star$ on manipulation and
above $1.26\,\mathcal{M}^\star$ on pursuit--evasion. At $q = 0.25$, the weakest
seed still returns 4.8 and 3.0 plans on Push and Slide, 9.6 and 7.5 times
IPOPT's count. On manipulation, the conclusions also hold across distinctness thresholds:
\method{}'s mean count is at least every baseline's at every threshold from
$0.5\,\delta_1$ to $8\,\delta_1$, and under the Fr\'echet distance. On pursuit--evasion, the ranking depends on the threshold. \method{} achieves the highest mean at $2\,\delta_1$ and $4\,\delta_1$. However, at the very fine $0.5\,\delta_1$ threshold, MPPI's tendency to generate closely spaced plans inflates its count (338 vs.\ 191). Conversely, at the extreme $8\,\delta_1$ threshold, where counts roughly correspond to the number of homotopy classes, IPOPT finds slightly more (1.5 vs.\ 1.3).

\subsection{Output Over Time at Equal Budgets}
\label{sec:anytime}

\begin{figure}[t]
\centering
\includegraphics[width=0.8\columnwidth]{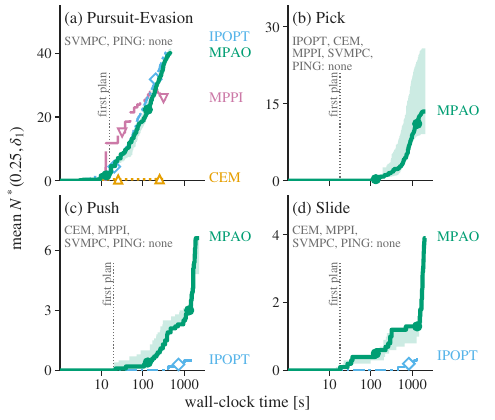}
\caption{\textbf{Distinct plans over time, at equal budgets.} Mean
$N^\star(0.25, \delta_1)$ over the 10 test instances against wall-clock time
(log scale).}
\label{fig:anytime}
\end{figure}

Fig.~\ref{fig:anytime} shows each method's plan set over time. Every baseline was run for a duration equal to \method{}'s search time on each instance. Furthermore, \method{}'s recorded times include its certificates (which account for a quarter to a third of its total wall-clock time), making this comparison strictly conservative for \method{}. On manipulation,
CEM, MPPI and SVMPC never return a plan within 25\% of $\mathcal{M}^\star$, and
IPOPT's first such plan appears after 344\,s on Push and 447\,s on Slide, ending at 0.5
and 0.4 plans per instance. \method{} overtakes IPOPT's final counts after just 169\,s and 128\,s, roughly a tenth of the total budget, and keeps growing to 6.6 and 3.9 plans per instance. On the Pick task, where no baseline finds a qualifying plan, \method{} finds its first after 104\,s and eventually reaches 13.5. On pursuit--evasion, IPOPT and MPPI start
faster (a first plan within the bound after 4\,s, and 12 plans after 13\,s), but \method{}
finishes level with IPOPT (39.9 vs.\ 40.3) and ahead of MPPI (25.8). Because its archive is strictly additive, \method{} can be terminated at any time to return the packing of all plans found so far. On manipulation tasks, its plan count is still rising when the budget expires.

\subsection{Ablation Study}
\label{sec:design}

\begin{figure}[t]
\centering
\includegraphics[width=0.6\columnwidth]{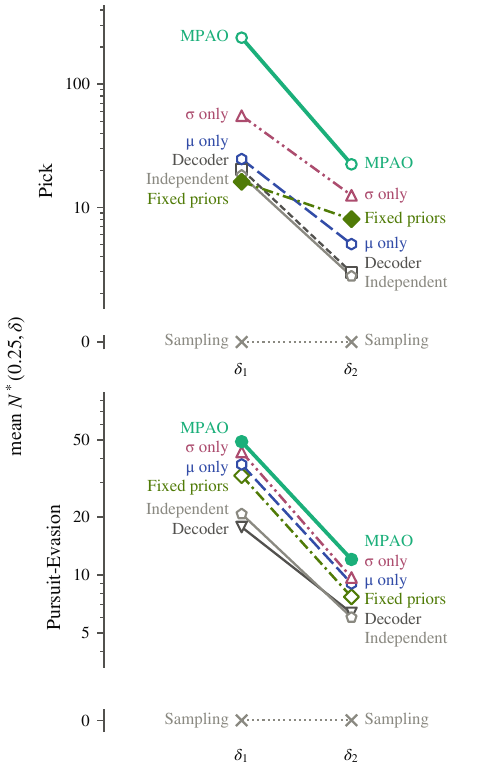}
\caption{\textbf{The full method is best in every setting.}
$N^\star(0.25, \delta)$ (log scale) at $\delta_1$ and $\delta_2$ for \method{}
and six degraded variants on two held-out instances.}
\label{fig:ablation}
\end{figure}

Fig.~\ref{fig:ablation} isolates \method{}'s components on one held-out instance each of Pick, where its lead is largest, and pursuit--evasion, where the baselines are strongest. To isolate the impact of each part, we evaluated six ablated variants. Three variants test the adaptation rules by removing replacement (\emph{$\sigma$ only}), removing widening (\emph{$\mu$ only}), or removing both (\emph{Fixed priors}). The remaining three alter the core architecture: using one global decoder without families or an archive (\emph{Decoder}), using one independent decoder per member (\emph{Independent}), or removing training entirely (\emph{Sampling}). The full \method{} performs best across all four settings. On Pick, it yields 4.3 and 1.8 times the plan count of the next best variant at $\delta_1$ and $\delta_2$, respectively. Furthermore, removing either adaptation rule consistently lowers the count in every setting. These rules prove highly complementary. On Pick, widening alone raises $N^\star(0.25, \delta_1)$ from 16 to 55, and replacement alone raises it to 25. However, combining both rules surges the count to 238. Notably, at the stricter $\delta_2$ threshold, replacement only improves performance when paired with widening. Finally, untrained \emph{Sampling} never yields plans within the cost bound. Meanwhile, both the \emph{Independent} and \emph{Decoder} variants return only 18 to 21 plans at $\delta_1$. Compared to the full \method{}, this requires 12-13x fewer plans on Pick and 2.4-2.8x fewer on pursuit-evasion.

\section{Limitations}
\label{sec:discussion}

\method{}'s design trades some capabilities for its strengths, which points to
directions for future work. First, it favours distinct plans over many plans
near the best known cost: once a member finds a plan, its slot is redrawn so
the search moves on, instead of refining that plan further. This explains why
IPOPT and MPPI, which locally refine around their best solutions, return more
plans within 10\% of $\mathcal{M}^\star$ on pursuit--evasion
(Sec.~\ref{sec:main-results}). Although \method{} reaches every route that
IPOPT finds, its plans on the second route cost more than
$1.25\,\mathcal{M}^\star$. In the future, polishing our archived plans with a
local optimizer could likely bring them within this tighter bound. Second, a
run takes 6 to 38 minutes per instance on a laptop GPU, and on Push and Slide
most plans within the bound appear in the second half of the run, as training
refines the shifts. \method{} therefore suits
precomputing plan sets offline rather than replanning inside a control loop.
Third, $N^\star$ counts plans that differ by at least $\delta$, which does not
inherently mean they represent different high-level strategies. At our thresholds,
\method{}'s counted plans span an average of 1.4 to 3.2 contact strategies per
manipulation instance, while on pursuit--evasion they mostly share a single route.
Therefore, metrics that explicitly evaluate discrete strategies or topological
routes would serve as a complement to $N^\star$. 

\section{Conclusion}
\label{sec:conclusion}

We framed continuous-time multi-plan planning as finding large $(q, \delta)$-plan sets, measured by the $N^\star(q, \delta)$ metric. To solve this, we presented \method{}, which samples families of base trajectories that satisfy the boundary conditions by construction, and shifts them with decoders shared within each family. Crucially, the algorithm archives every validated plan and adapts its families as this archive grows. Under equal budgets, \method{} is the only method that solves all 40 instances. On pursuit--evasion, it matches the strongest baseline overall, and takes the lead at the stricter distinctness threshold or when timing is ignored. On the Push and Slide tasks, it returns approximately ten times as many distinct, low-cost plans as the strongest baseline, and it is the only method that finds any plan within 25\% of the best known cost on Pick. Future work includes refining these plans closer to the best known cost and executing them on physical hardware.

\bibliography{aaai2027}

\end{document}